\documentclass[twocolumn]{article}

\usepackage{url}       
\usepackage{graphicx}  
\usepackage{float}

\usepackage{makecell}
\usepackage{arydshln}
\usepackage[dvipsnames,table,xcdraw]{xcolor}

\usepackage{amsmath}
\usepackage{amssymb}
\usepackage{hyperref}       
\usepackage{url}
\usepackage{float}

\usepackage{booktabs}

\usepackage{easylist}
\usepackage{soul}
\usepackage{bm}
\usepackage{colortbl}

\usepackage{charter}
\usepackage{eulervm}

\newcommand{\N}{{\mathbb N}}

\newcommand{\Z}{{\mathbb Z}}

\newcommand{\B}{{\mathbb B}}

\newcommand{\p}{\;\text{.}}

\newcommand{\I}{{\cal I}}
\newcommand{\tab}{\hspace*{5mm}}

\newcommand{\h}{\hspace{0.7mm}}

\definecolor{myred}{HTML}{C82506}
\definecolor{myblue}{HTML}{0365C0}
\definecolor{mygreen}{HTML}{00882B}
\definecolor{mydarkgreen}{HTML}{0B5D18}
\definecolor{myorange}{HTML}{DE6A10}
\definecolor{mypurple}{HTML}{773F9B}
\definecolor{myyellow}{HTML}{DCBD23}
\definecolor{mygray}{HTML}{999999}

\newcommand{\rc}[1]{{\textcolor{myred}{#1}}}
\newcommand{\gc}[1]{{\textcolor{mygreen}{#1}}}

\newcommand{\bc}[1]{{\textcolor{myblue}{#1}}}

\newcommand{\G}{\bc{\cal G}}
\newcommand{\cS}{\bc{\cal S}}

\floatstyle{ruled}
\newfloat{pseudo}{h}{lop}
\floatname{pseudo}{Algorithm}

\begin{document}

\title{Finding Motifs in Knowledge Graphs using Compression}

\author{Peter Bloem, Vrije Universiteit Amsterdam\\ 
\small vu@peterbloem.nl, ORCID: 0000-0002-0189-5817}

\maketitle

\begin{abstract}
\noindent We introduce a method to find \emph{network motifs} in knowledge graphs. Network motifs are useful patterns or meaningful subunits of the graph that recur frequently. We extend the common definition of a network motif to coincide with a \emph{basic graph pattern}. We introduce an approach, inspired by recent work for simple graphs, to induce these from a given knowledge graph, and show that the motifs found reflect the basic structure of the graph. Specifically, we show that in random graphs, no motifs are found, and that when we insert a motif artificially, it can be detected. Finally, we show the results of motif induction on three real-world knowledge graphs.
\end{abstract}

\section{Introduction}
\noindent \emph{Knowledge graphs} are an extremely versatile and flexible data model. They allow knowledge to be encoded without a predefined format and they are extremely robust in the face of missing data. This versatility comes at a price. For a given knowledge graph, it can be difficult to see the forest for the trees: how is the graph structured at the lowest level? What kind of things can I ask of what types of entities? What are small, recurring patterns that might represent a novel insight into the data? Answering these questions could benefit problem domains like graph simplification, graph navigation and schema induction.

In the domain of unlabeled simple graphs, \emph{network motifs} \cite{milo2002network} were introduced as a tool to provide insight into local graph structure. Network motifs are small subgraphs whose frequency in the graph is unexpected with respect to a \emph{null model}. 


\begin{figure*}[bth]
  \centering
    \includegraphics[width=\textwidth]{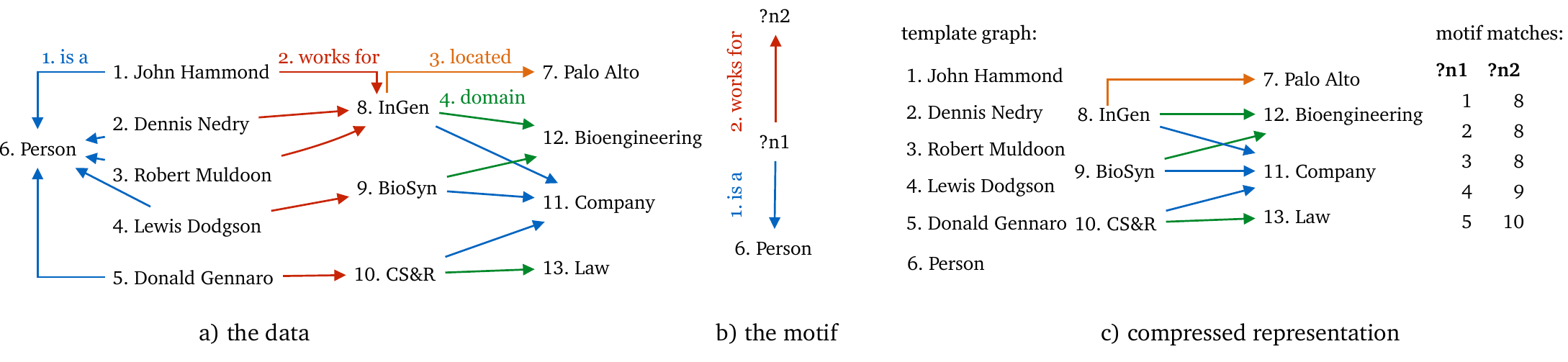}
    \caption{An example of the principle behind our motif code. a) A basic knowledge graph. We consider only the integer indices of the nodes and relations. Labels are included only for readability. b) A motif that occurs frequently. c) A compressed representation; we remove all edges that are part of an occurrence of the motif and store separately which nodes match the motif. Together with a the motif, this allows us to reconstruct the data.}
    \label{figure:example}
  \label{figure:codes}
\end{figure*}

Unfortunately, estimating this probability usually requires repeating the subgraph count on many samples from the null model. To avoid this costly operation, \cite{bloem2017large} introduces  an alternative method, using \emph{compression} as a heuristic for motif relevance: the better a motif compresses the data, the more likely it is to be meaningful. 



In this paper, we extend this compression-based motif analysis to \emph{knowledge graphs}. For the purposes of this research we define knowledge graphs as labeled, directed multigraphs. Nodes are uniquely labeled with entity identifier, and links are non-uniquely labeled with relations.  We extend the definition of a motif to that of a \emph{basic graph pattern}: a small graph labeled with both variables and explicit entitites and relations. A pattern matches if the variables can be replaced with specific values from the graph so that the pattern becomes a subgraph as a result. 

The intuition behind our method is that we can use graph patterns to \emph{compress the graph}: we store the pattern, its instances, and the remainder of the graph. The better this representation compresses the graph, the more relevant the pattern. Figure~\ref{figure:example} illustrates the principle. In Section~\ref{section:preliminaries}, we justify this intuition more formally.

We perform several experiments to show that our method returns meaningful subgraphs. First we test the intuition that a random graph should contain no motifs. We also show that when we artificially insert motifs into a random graph, we can then detect these as motifs. Finally, we show the results of motif analysis on three real-world knowledge graphs, compared to the baseline of selecting the most frequent graph patterns.

All code and datasets used in this paper are available online.\footnotemark

\footnotetext{\url{https://github.com/MaestroGraph/motive-rdf}}

\paragraph{Related Work}

Network motifs for unlabeled simple graphs were introduced in \cite{milo2002network}. A more comprehensive overview of the related literature can be found in \cite{bloem2017large}. In \cite{bloem2017large}, the principle of Minimum Description Length (MDL) was first connected to motif analysis. However, the idea had earlier been exploited for detecting meaningful subgraphs in the SUBDUE algorithm \cite{cook1994substructure}.

A few other methods have been proposed for inducing the structure of a given knowledge graph in terms of subgraphs. In \cite{pham2015deriving}, the authors use the principle of characteristic sets to characterize a knowledge graph in terms of the star patterns it contains. In \cite{pham2016exploiting}, they show that the majority of the LOD cloud can be efficiently described using such principles, showing the highly tabular structure of many knowledge graphs. In \cite{volker2011statistical}, association rule mining is used to induce basic patterns in the graph. 

To the best of our knowledge, ours is the first method presented that can potentially induce any basic graph pattern.

\subsection{Preliminaries}

\label{section:preliminaries}

\paragraph{Minimum Description Length}
Our method is based on the MDL principle: we should favour models that compress the data. We will show briefly how this intuition can be made mathematically precise. For more details, we refer the reader to \cite{grunwald2007minimum} for MDL in general, and to \cite{bloem2018tutorial}, for a more extensive discussion these principles in the domain of graph analysis.
 
Let $\B$ be the set of all finite-length binary strings. We use $|b|$ to represent the length of $b \in \B$. Let $\log(x) = \log_2(x)$. A \emph{code} for a set of objects $\cal X$  is an injective function $f: {\cal X} \to \B$. All codes in this paper are \emph{prefix-free}: no code word is the prefix of another. We will denote a \emph{codelength function} with the letter $L$, ie. $\rc{L}(x) = |f(x)|$. We commonly compute $\rc{L}(x)$ directly, without first computing $f(x)$.

There is a strong relation between codes and probability distributions: for each probability distribution $\rc{p}$ on $\cal X$, there exists a prefix-free code $\rc{L}$ such that for all $x \in \cal X$: $- \log \rc{p}(x) \leq \rc{L}(x) < -\log \rc{p}(x) + 1$. Inversely, for every prefix-free code $L$ for $\cal X$, there exists a probability distribution $\rc{p}$ such that for all $x \in \cal X$: $\rc{p}(x) = 2^{-\rc{L}(x)}$. For proofs, see \cite[Section~3.2.1]{grunwald2007minimum} or \cite[Theorem~5.2.1]{cover2006elements}.



\paragraph{Relevance testing} We will use the MDL principle to perform a hypothesis test. Assume we have some data $x \in \B$ and a null hypothesis stating that it was sampled from distribution $\rc{p^\text{null}}$ (with corresponding code $\rc{L^\text{null}}$). A simple but crucial result, known as the \emph{no-hypercompression inequality} \cite[p103]{grunwald2007minimum} tells us that the probability of sampling any data $x$ from $\rc{p^\text{null}}$ that can be described in less than $\rc{L^\text{null}}(x) - k$ or more bits, \emph{using \bc{any code}} is less than $2^{-k}$.
Thus, we can reject the hypothesis that the data was sampled from $\rc{p^\text{null}}$ by designing \bc{an alternative code} $\bc{L^\text{alt}}$ which compresses the data better than $\rc{L^\text{null}}$ by, say, 10 bits ($\rc{L^\text{null}}(x) - \bc{L^\text{alt}}(x) \geq 10$) and rejecting the null hypothesis with confidence $2^{-10}$. For a longer, more intuitive explanation of this principle in pattern induction, we refer the reader to \cite{bloem2018tutorial}.

Note that when we use this procedure to find motifs, we are not providing statistical evidence for the hypothesis that the motif is ``correct'' \cite[Section~6.1]{bloem2017large}. We are simply using the principle of hypothesis testing as a \emph{heuristic} for pattern mining. The only assertion we are proving (in a statistical sense) is that the data did not come from the null model.

\paragraph{Common codes}
In the construction of our graph codes, we require some simpler codes as building blocks. First, when we store any positive integer $n$, we do so with the code corresponding to the distribution $p^\N(n) = 1/(n(n+1))$, and denote it $L^\N(n)$. For nonnegative numbers we add 1 to the argument. For the full range of integers ($L^\Z$), we add an extra bit for the sign, and then use the first code for negative integers and the second for positive ones.

We will often need to encode \emph{sequences} of integers as well. These will be highly skewed, with only a subset of integers occurring frequently, and others occurring infrequently or not at all. As noted in \cite{de2016names} a code based on the Pitman-Yor model \cite{pitman1997two} is very effective in such situations. Let $\rc{S} = \langle \gc{S_1}, ..., \gc{S_n}\rangle$ be a sequence of integers of length $n$. We first store the set of its members $m(\rc{S})$ (the vocabulary) in the order in which they occur: we store $n$ and the first member using $L^\N$ and then store each subsequent member by encoding the distance to the previous member using $L^\Z$. Having encoded the members of $\rc{S}$ we can store the sequence itself using the Pitman-Yor model as follows.

Let $f(\gc{A}, \rc{B})$ be the frequency of symbol $\gc{A}$ in sequence $\gc{B}$. We then store the complete sequence using the code corresponding to the following distribution:
\begin{align*}
p(\rc{S}) &= \prod_{i \in [1,k]} p(\gc{S_i}\mid \rc{S_{1:i-1}})\\
& \text{with}\;\;
p(\gc{A} \mid \rc{B}) = 
\begin{cases}
\frac{\bc{\alpha} - \bc{d} |m(\rc{B})|}{|m(\rc{B})| + \bc{\alpha}} & \text{if } f(\gc{A}, \rc{B}) = 0 \\ 
\frac{f(\gc{A}, \rc{B}) - \bc{d}}{|m(\rc{B})| + \bc{\alpha}} & \text{otherwise}
\end{cases}
\end{align*}

See \cite{de2016names} for a more intuitive explanation. In all experiments we use $\bc{\alpha} = 0.5$, $\bc{d}=0.1$. We will refer to the total resulting codelength as $L^{PY}(S)$.

\section{Method}

We will first give a precise definition of a knowledge graph as used in this paper. We will then describe the null model which is used both as a point of comparison in our hypothesis test, and within the motif code to compress the remainder of the graph. Next, we describe how to compress a graph using a given motif, and a set of instances. Finally, we will describe how to search for likely motifs using simulated annealing.

We analyse the \emph{structure} of knowledge graphs only, ignoring any meaning in relation to other graphs, encoded in the content of names or literals, or from ontology languages. Specifically, we model a knowledge graph as a multigraph with nodes and edges labeled with integers that map to entities and relations. This mapping is stored, but only the integer-labeled graph is modelled.\footnotemark
\footnotetext{For practitioners this restriction is not noticeable, as the indices can simply be mapped back to the original strings when the found motifs are presented.}

A \textbf{knowledge graph} $G$, is a tuple $G = (\gc{v}_G, \rc{r}_G, \bc{E}_G)$. $\gc{v}_G \in \N$ is the number of nodes in the graph, and $\rc{r}_G \in \N$ is the number of relations. We define the nodeset of $G$ as $\rc{V}_G = \{0, \ldots, \gc{v}_G-1\}$ and the relation-set as $\rc{R}_G = \{0, \ldots, \rc{r}_G\}$. The \emph{tripleset} $\bc{E}_G \subseteq \gc{V}_G \times \rc{R} \times \gc{V}_G$ determines the edges of the graph and their labels: each triple $(\gc{s}, \rc{p}, \gc{o}) \in \bc{E}_G$ encodes the subject node $\gc{s}$, the object node $\gc{o}$ and the \emph{predicate} or \emph{relation} $\rc{p}$ of an edge in the graph. 

This definition is compatible with RDF data. We interpret literals as nodes, considered the same node if they are expressed by the same string.


A \textbf{pattern} $M$ for graph $G$ is a tuple $(\gc{V}_M, \rc{R}_M, G, \bc{E}_M)$. Let $\gc{v'}_M$ and $\rc{r'}_M$ indicate the number of variable nodes and variable links in $M$ respectively, then $\gc{V}_M \subseteq \{-\gc{v'}_M, \ldots, \gc{v}_G-1\}$ and $\rc{R}_M \subseteq \{-(\rc{r'}_M+\gc{v'}_M), \ldots,-\gc{v'}_M, 0,\ldots, \rc{r}_G-1\}$, with $\bc{E}_M \subset \gc{V}_M \times \rc{R}_M \times \gc{V}_M$ representing the edges as before. That is; nodes in a pattern can be labeled either with nonnegative integers referring to $G$'s nodes or with negative integers representing a variable node, and similar for relations. The negative integers are always contiguous within a single pattern, with the highest representing the node labels and the lowest representing the edge labels

An \textbf{instance} for pattern $M$ in graph $G$ is a pair of sequences of integers: $I = (\gc{I^n}, \rc{I^r})$. $\gc{I^n}$ is a sequence of distinct integers of length $\gc{v}_M$. $\rc{I^r}$ is a sequence of non-distinct integers of length $\rc{r}_M$. For each edge $(\gc{s}, \rc{p}, \gc{o}) \in \bc{E}_M$ with any or all of $\gc{s}$, $\rc{p}$ and $\gc{o}$ negative, there is a corresponding link in $\bc{E}_G$ with a negative $\rc{s}$ replaced by $\gc{I}^\gc{n}_{-\gc{s}}$, a negative $\gc{o}$ replaced by $\gc{I}^\gc{n}_{-\gc{o}}$, and a negative $\rc{p}$ replaced by $\rc{I}^\rc{r}_{-\rc{p} - \gc{v'}_M}$. Put simply: for a pattern to match, variable edges marked with the same negative integer, must map to the same relation in order for the pattern to match, but variable links labeled with different negative integers \emph{may} map to the same relation. Variable nodes are always labeled distinctly and may never map to the same node in $G$. An instance describes a subgraph of $G$ that \emph{matches} the pattern $M$. Each edge in the motif may only match one edge in the graph. In other words, the occurrence of the motif in the graph must have as many edges as the motif itself.\footnotemark

\footnotetext{In this aspect our definition differs from the SPARQL Basic Graph Pattern. Patterns for which this distinction is relevant are rare, and patterns returned by our method are still compatible with SPARQL.}

\subsection{Null model}

For a proper hypothesis test, we must compare the compression achieved by our motif code to the compression under a general model for knowledge graphs: a null model. 

The most common null model in classical motif analysis is the degree-sequence model (also known as the configuration model \cite{newman2010networks}): a uniform distribution over all graphs with a particular degree sequence. We extend this to knowledge graphs by also including the degree of each relation: that is, degree of a relation is the frequency with which it occurs in the tripleset. Let a \emph{degree sequence} $D$ of length $n$ be a triple of three integer sequences: $(\gc{D^\text{in}}, \rc{D^\text{rel}}, \gc{D^\text{out}})$. If $D$ is the degree sequence of a graph, then node $i$ has $\gc{D}^\gc{\text{in}}_i$ incoming links,  $\gc{D}^\gc{\text{out}}_i$ outgoing links and for each relation $r$, there are $\rc{D}^\rc{\text{rel}}_r$ triples.

Let $\bc{\G}_D$ be the set of all graphs with degree sequence $D$. Then the degree-sequence model can be expressed simply as
\[
p^\text{DS}(G) = \frac{1}{|\G_D|}
\]
for any $G$ that satisfies $D$ and $p(G) = 0$ otherwise. Unfortunately, there is no efficient way to compute $|\G_D|$ and even approximations tend to be costly for large graphs. Following the approach in \cite{bloem2017large}, we define a fast approximation to the configuration model, which works well in practice for motif detection. 

We can describe a knowledge graph by three length-$m$ integer sequences: $\gc{S}$, $\rc{P}$, $\gc{O}$, such that $\{(\gc{S}_j, \rc{P}_j, \gc{O}_j)\}_j$ is the graph's tripleset. If the graph satisfies degree sequence $D$, then we know that $S$ should contain node $j$ $\gc{D}^\gc{\text{out}}_j$ times, $\rc{P}$ should contain relation $r$ $\rc{D}^\rc{\text{rel}}_r$ times and $\gc{O}$ should contain node $j$ $\gc{D}^\gc{\text{in}}_j$ times.  Let ${\cal S}_D$ be the set of all such triples of integer sequences satisfying $D$. We have 
\[
|{\cal S}_D| =
 {m \choose {\gc{D}_1^\gc{\text{out}}, \ldots, \gc{D}_n^\gc{\text{out}}} }
 {m \choose {\rc{D}_1^\rc{\text{rel}}, \ldots, \rc{D}_{|R_G|}^\rc{\text{rel}}} }
 {m \choose {\gc{D}_1^\gc{\text{in}}, \ldots, \gc{D}_n^\gc{\text{in}}} } \text{.}
\]

While every member of $\bc{\cal S}_D$ represents a valid graph satisfying $D$, many graphs are represented multiple times. Firstly, many elements of $\bc{\cal S}_D$ contain the same link multiple times. We call the set without these elements $\bc{\cal S'}_D \subset \bc{\cal S}_D$. Secondly the links of the graph are listed in arbitrary order; if we apply the same permutation to all three lists $\gc{S}$, $\rc{P}$ and $\gc{O}$, we get a new representation of the same graph. Since we know that any element in $\bc{\cal S}'_D$ contains only unique triples, we know that each graph is present exactly $m!$ times. This gives us
\[
|\bc{\G}_D| = |\bc{\cal S}'_D| \frac{1}{m!} \leq  |\bc{\cal S}_D| \frac{1}{m!} \text{.}
\]

We can thus use 
\[
p^\text{EL}_D(G) =  \frac{m!}{|\bc{\cal S}_D|} \leq p^\text{DS}(G) \text{.}
\]

Filling in the definition of the multinomial coefficient, and rewriting, we get a codelength of:
\begin{align*}
- \log p^\text{EL}_D(G) =&\;\; 2 \log(m!) - \sum_i \log(\gc{D}_i^\gc{\text{in}}!)\\
&\;- \sum_i \log(\rc{D}_i^\rc{\text{rel}}!) - \sum_i \log(\gc{D}_i^\gc{\text{out}}!)
\end{align*}

as an approximation for the DS model. We call this the edgelist (EL) model. It gives a probability that always lower-bounds the configuration model, since it affords some probability mass to graphs that cannot exist. Experiments in the classical motif setting have shown that the EL model is an acceptable proxy for the DS model \cite{bloem2017large}, especially considering the extra scalability it affords.

%
 \paragraph{Encoding D} In order to encode a graph with $L^\text{EL}_D$, we must first encode $D$.\footnotemark~For each of the three sequences $D'$ in $D$ we use the following model:
\[
 p(D') = \prod_i q^{\N}(D'_i)\;\;\;\;\;
 L(D') = - \sum_i \log q^{\N}(D'_i)
\]

\footnotetext{Or, equivalently, to make $p^\text{EL}$ a complete distribution on all graphs, we must provide it with a prior on $D$.}

where $1^{\N}$ is any distribution on the natural numbers. This is an optimal encoding for $D$ assuming that its members are independently drawn from $q^{\N}$. When we use $p^\text{EL}$ as the null model, we use the data distribution for $q^{\N}$ to ensure that we have a lower bound to the optimal code-lenngth (in essence, we cheat in favor of the null model,giving it a slightly lower than optimal codelength). When we use $p^\text{EL}$ as part of the motif code, we must use a fair encoding, so we use the Pitman-Yor code to store each sequence in $D$.

In the design of our method, we will constantly aim to find a trade-off between completeness and efficiency that allows the method to scale to very large graphs. Specifically, when we economize, we will only do so in a way that makes the hypothesis test \emph{more conservative}.

\subsection{Motif code}

Having defined our representation of a knowledge graph, and a general null model for compressing such knowledge graphs, we can now define how we use a given pattern (together with its instances) to compress a dataset. 

We will assume that a target pattern $M$ is given for the data $G$ and that we have a set of instances $\I$ of $M$ in $G$. Moreover, we require that all instances in $\I$ are mutually disjoint: no two subgraphs defined by a member of $\I$ may share an edge, but nodes may be shared. Given this information, we will define a motif code that will help us determine whether or not $M$ is a likely motif for $G$. In section Section~\ref{section:search}, we detail a method to search for pairs $(M, \I)$ to pass to the motif code.

As described above, we can perform our relevance test with any compression method which exploits the pattern $M$ and its instances $\I$ to store the graph efficiently. The better our method, the more motifs we will find. Note that there is no need for our code to be optimal in any sense. We know that we will not find all motifs that exist, and we will not use them optimally to represent the graph, but the test is still valid. This also means that we are free to trade off compression performance against efficiency of computation.

We store the graph by encoding various aspects, one after the other. The information in all of these together is sufficient to reconstruct the graph. Note that everything is stored using prefix-free codes, so that we can simply concatenate the codewords we get for each aspect, to get a codeword for the whole graph.

We also assume that we are given a code $L^\text{base}$ for generic knowledge graphs (in practice, this will be the null model, although the motif code is valid for any base code).

We store, in order:

\begin{description}
 \item[the graph dimensions] We first store $\gc{v}_G$, $\rc{r}_G$ and $|\bc{E}_G|$ using the generic code $L^{\N}(\cdot)$. 
 \item[the pattern] We store the structure of the pattern using the base code, and its labels as a sequence using the Pitman-Yor code.
 \item[the template] This is the graph, minus all links occurring in instances of $M$. Let $\bc{E}_G'$ be $\bc{E}_G$ minus any link occurring in any member of $\cal I$. We then store $(\gc{v}_G, \rc{r}_G, \bc{E'}_G)$ using $L^\text{base}(\cdot)$.
 \item[the instances] To store the instances, we view the connections between the nodes made by motifs as a hypergraph, and we extend the EL code to store it. The details are given below.
\end{description}

The precise computation of the codelength is given in Algorithm~\ref{algorithm:motif-code}. 

\begin{pseudo}[tb]
\caption{The motif code $L^\text{motif}(G ; M, {\cal I}, L^\text{base})$. Note that the nodes and relations of the graph are integers.}
\label{algorithm:motif-code}
{ \small
\textbf{function} $\text{codelength}(G; M, {\cal I}, L^\text{base})$:\\
\tab\tab a graph $G$, a pattern $M$\\ 
\tab\tab instances $\cal I$ of $M$ in $G$, a code $L^\text{base}$.\\
\\
$b_\text{dim} \leftarrow L^\N(\gc{v}_G) + L^\N(\rc{r}_G) + L^\N(|\bc{E}_G|)$ \\

---\emph{Turn the pattern into a normal knowledge graph}\\
$\bc{E}_{M'} \leftarrow$ the edges of $M$ with positive integer labels \\
$M' \leftarrow (\gc{v}_{M}, \rc{r}_{M}, \bc{E}_{M'})$ \\
$S_M \leftarrow$ the labels of $M$ in canonical order \\
$b_\text{pattern} \leftarrow L^\text{base}(M') + L^{PY}(S_M) $\\

---\emph{Store the template graph}\\
$\bc{E'}_G \leftarrow \bc{E}_G - \cup_{\I \in {\cal I}} \text{triples}(I)$ \\
$b_\text{template} \leftarrow L_\text{base}((\gc{v}_G, \rc{r}_G, \bc{E'}_G))$\\

$b_\text{instances} \leftarrow -\log p_M(\I) + \sum_{D \in D^{\I}} L^{PY}(D)$\\

\textbf{return} $b_\text{dim} + b_\text{pattern} + b_\text{template} + b_\text{instances}$\\
}
\end{pseudo} 

\paragraph{Encoding motif instances} To encode a list of instances $\I$ of a given pattern $M$, we generalize  the idea of the edgelist model described above. 

 
 To generalize this notion to arbitrary patterns, to be defined for a given template graph, we define the \emph{degree constraint} $D^\I$ of a list of instances for a given pattern as follows: for each variable node $\gc{i}$ in the pattern, the degree constraint provides an integer sequence $\gc{D^i}$ of length $\gc{v}_G$, indicating how often each node in the completed knowledge graph takes that position in the pattern. Similarly, for each variable edge $\rc{j}$ in the pattern, the degree constraint provides an integer sequence $\rc{C^j}$ of length $\rc{r}_G$ indicating for each relation how often it takes that position in the pattern.
 
 We store these sequences in the same manner as the degree sequence of the template graph, using the Pitman-Yor code for each.
 
Given this information, all we need to do is describe which of the possible sequences of matches for this pattern satisfying the given degree constraint we are encoding. As with the configuration model, the ideal is a uniform code over all possible configurations, for which we will define an approximation. Given $\gc{w}$ variable nodes in a pattern, and $\rc{l}$ variable edges, we can define such a collection of instances using $\gc{w}+\rc{l}$ integer sequences: $\gc{N^1}, \ldots, \rc{N^n}, \rc{L^1}, \ldots, \rc{L^l}$, with the $t$-th instance defined by the integer tuple $(\rc{N}^\rc{1}_t, \ldots, \gc{N}^\gc{n}_t, \rc{L}^\rc{1}_t, \ldots, \rc{L}^\rc{l}_t)$. If this set of sequences satisfies the degree constraint, we know that node $q$ must occur $\gc{D}^\gc{i}_q$ times in sequence $\gc{N}^\gc{i}$, and similarly for the variable links. Let $\cS_\I$ be the set of all such integer sequences satisfying the  constraint. We follow the same logic as for the EL model. Let $k$ be the number of matches of the pattern. We have:
\begin{align*}
|\cS_\I| = &{k \choose \gc{D}^\gc{1}_1, \ldots, \gc{D}^\gc{1}_v}\times \ldots \times{k \choose \gc{D}^\gc{w}_1, \ldots, \gc{D}^\gc{w}_v}\times \\
&{k \choose \rc{C}^\rc{1}_1, \ldots, \rc{C}^\rc{1}_r}\times \ldots \times{k \choose \rc{C}^\rc{l}_1, \ldots, \rc{C}^\rc{l}_r}
\end{align*}

As before, this set is larger than the set we are interested in. First, each set of pattern matches is contained multiple times (once for each permutation) and second, not all elements are valid pattern matches (in some, a single triple may be represented by multiple instances). Let $\cS_\I'$ be the subset representing only valid matches, and let $\G_\I$ be the set of valid instances with permutations removed. As before, we have
\[
|\G_\I| = |\cS'_\I| \frac{1}{k!} \leq  |\cS_\I| \frac{1}{k!} \text{.}
\]
Which gives us the following distribution
\[
p_M(G) =  \frac{k!}{|\cS_D|} < \frac{1}{\G_D} \text{,}
\]
 with $-\log p_M(\I)$ as a code to store the instances. Rewriting as before, gives us a codelength of 
 \begin{align*}
 - \log p_M(G) =&\;\; (\gc{w}+\rc{l}-1)\log(k!) \\
  &\;-\sum_{\gc{j}\in [1,\gc{w}], i} \log(\gc{D}^\gc{j}_i!) -\sum_{\rc{j}\in [1,\rc{l}], i} \log(\rc{C}
  ^\rc{j}_i!)
 \end{align*}
 
Note that if we store a graph with the pattern \texttt{?n1 ?rel ?n2} we obtain an empty template graph, and this code reduces to the EL code, achieving the same codelength as the edgelist model, up to a small constant amount for storing the pattern.

For a given graph and pattern, we can simply find the complete list of instances using a graph pattern search. Since we require a slightly different semantics than standard graph pattern matchers, we adapt the DualIso algorithm \cite{saltz2014dualiso} for knowledge graph matching. Before computing the motif code, we prune the list of instances provided by this search iterating over the instances and removing any instance that produces a triple also produced by an earlier instance. To guard against rare patterns that produce long-running searches we terminate all searches after 5 seconds, returning  only those matches that were found within the time limit.

We express the strength of a motif by its log-factor: 
\[
L^\text{null}(G) - L^\text{motif}(G; M, \I, L^\text{base})\p
\] 
If this value is positive, the motif code compresses the graph better than the null model. If the log-factor is greater than 10 bits, it corresponds to a rejection of the null model at $p < 0.001$.

\subsection{Motif search}

\begin{figure}[tb]
  \centering
    \includegraphics[width=\linewidth]{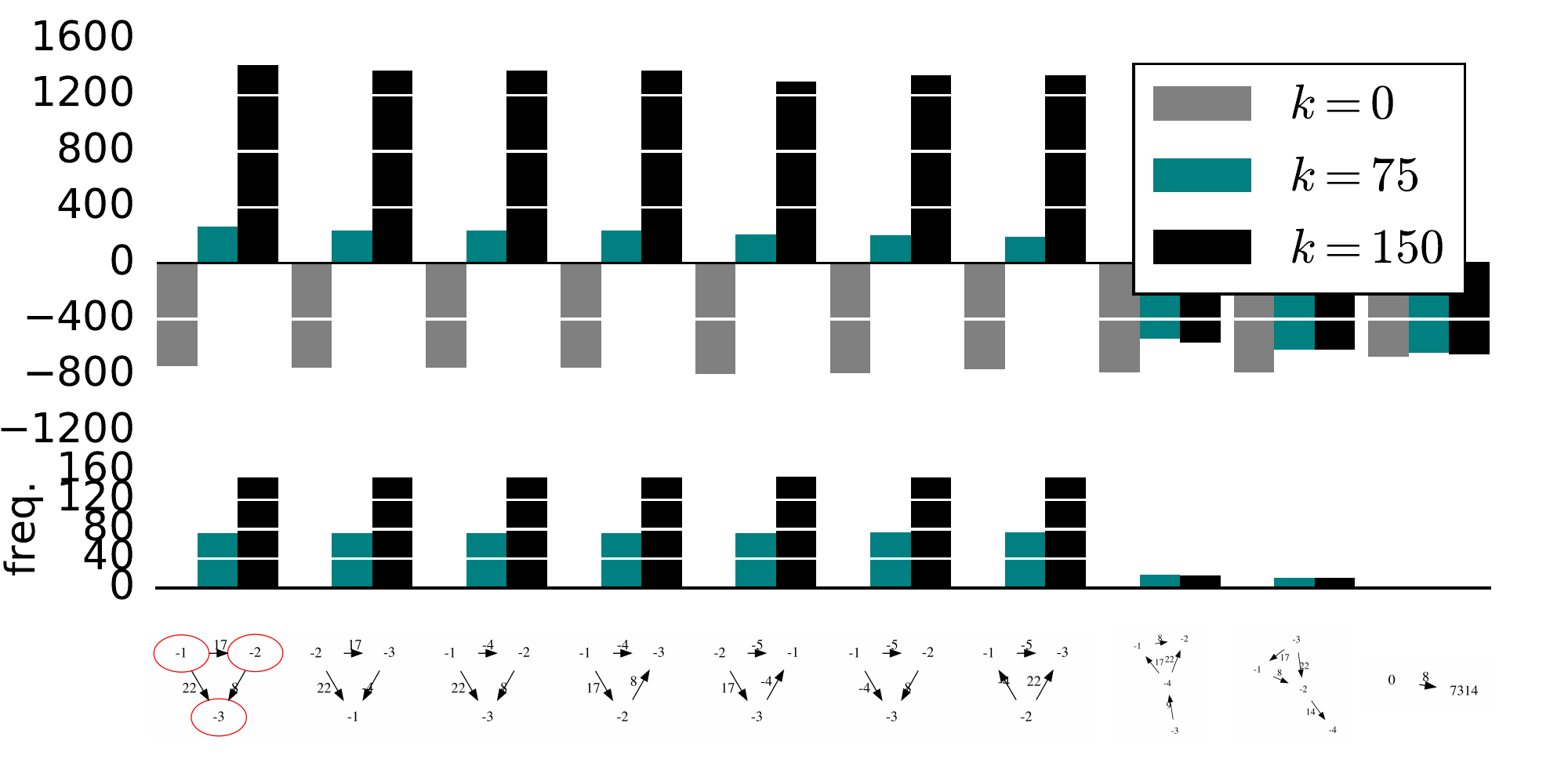}
    \caption{The result of the random graph experiment. We sort the motifs by their score in the $k=75$ experiment and plot their frequency and log-factor.}
    \label{figure:synthetic}
\end{figure}

\label{section:search}

Ultimately, we want to find any patterns that have a high log-factor for a given graph $G$. Since we can readily compute the log-factor for any given pattern, any black-box optimization algorithm can be used to search the space of all possible motifs. For the sake of simplicity, we will use basic simulated annealing: We start with a given pattern, and iterate by modifying the pattern in one of seven ways, chosen randomly. At each iteration, we search for instances of the pattern (limiting the time per search to 5 seconds) and compute the log-factor. If the log factor is better, we move to the new pattern, if it is worse, we move to the new pattern with probability 0.5. 

The starting pattern is always a single random triple from the graph, with its relation made a variable. We define seven possible transition from one pattern to another: \\
\textbf{Extend} Choose an instance of the pattern and an adjacent triple not part of the instance. Add the triple to the pattern.\\
\textbf{Make a node a variable} Choose a random constant node, and turn it in to a variable node.\\
\textbf{Make an edge a variable } Choose a random constant edge label, and turn it in to a variable (always introducing a new variable).\\
\textbf{Make a variable node constant} Choose a random variable node and turn it into a constant. Take the value from a random instance.\\
\textbf{Make a variable edge constant} Choose a random variable edge and turn it into a constant. Take the value from a random instance.\\
\textbf{Remove an edge} Remove a random edge from the pattern, ensuring that it stays connected.\\
\textbf{Couple} Take two distinct edge variables, which for at least one instance hold the same value and turn them into a single variable.

All transitions are equally likely. If the transition cannot be made (for instance, there are no constant nodes to make variable) or if the resultant pattern is in some way invalid, we sample a new transition.

Once a new pattern has been sampled, we compare its codelength under the motif model to that of the previous sample. If the codelength is lower, we continue with the new pattern. If the codelength is longer, we continue with the new sample with probability $\alpha$ or return to the previous pattern otherwise. We use $\alpha = 0.5$ in all experiments.

We store all encountered patterns and their scores. In order to exploit all available processor cores, we run several searches in parallel. We take the top 1000 patterns from each and sort them by motif codelength. Variables are re-ordered to a canonical ordering using the Nauty algorithm \cite{mckay1981practical}, so that isomorphic patterns are not tested twice.
 
 \begin{figure}[tb]
  \centering
    \includegraphics[width=0.9\linewidth]{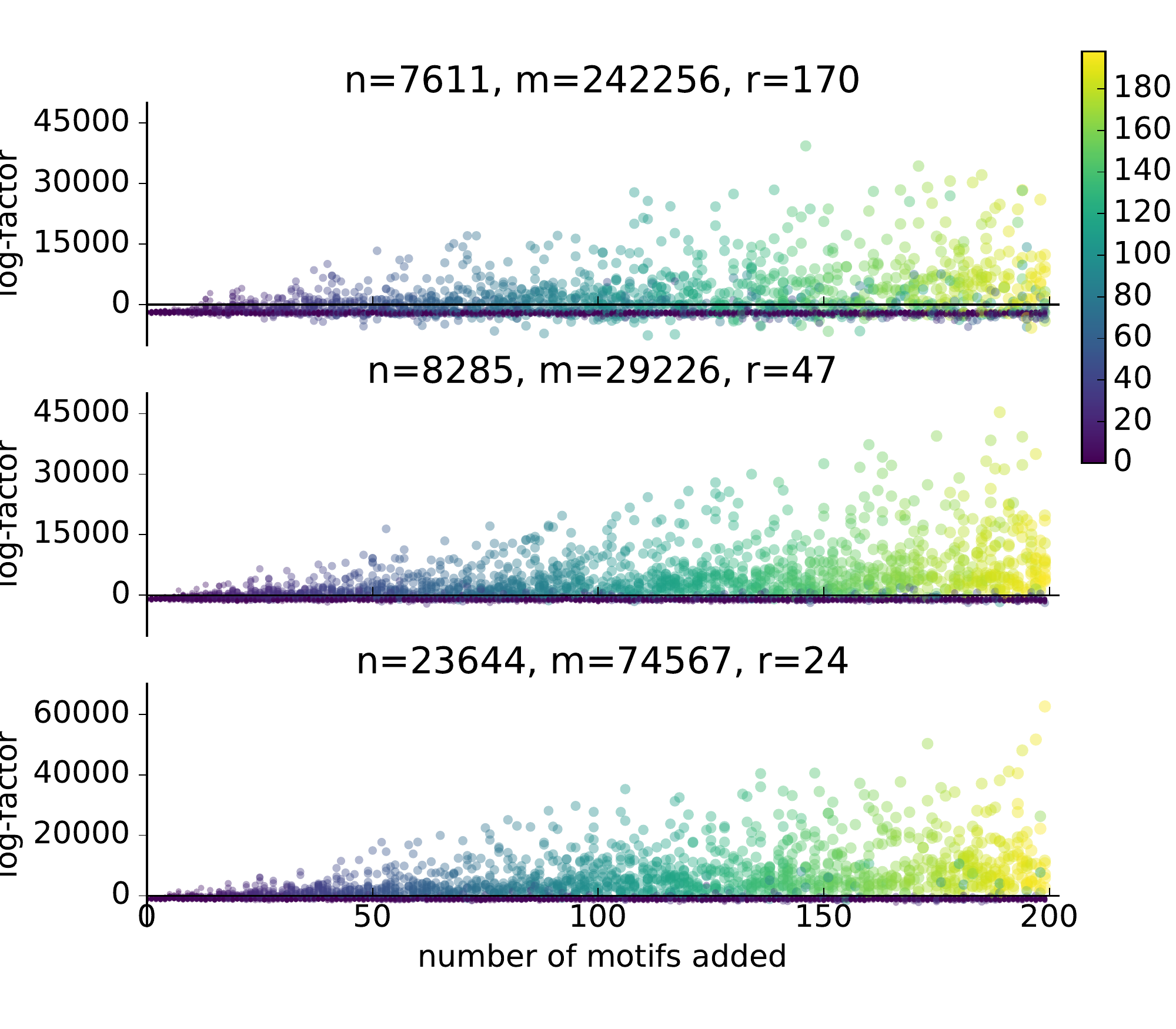}
    \caption{The result of the repeated random graph experiment. Color and size show the number of matches of the pattern after pruning. Plot titles show the graph dimensions before adding instances.}
    \label{figure:synthrep}
\end{figure}
\section{Experiments}

\paragraph{Random graphs} To validate the method, we first test it on random graphs. The aim is to test two requirements of a succesful pattern mining method:
\begin{itemize}
\item In a fully random graph, there should be no motifs, and we do not expect a motif code to outperform the null model.
\item If we insert a small number of instances of a particular pattern into the graph manually, these should be recognized as motifs. 
\end{itemize}

We sample a directed graph with a given number of nodes $n$ and edges $m$, with no self-connections and multiple edges (that is, we sample from from the $G(n, m)$ Erd\H{o}s-Renyi model). We then label the nodes uniformly at random with one of the relations in $0, \ldots, \rc{r}$. To make the dimensions realistic we base them on those of MUTAG dataset used in the next section.

We then take one randomly chosen pattern, and insert $\bc{k}=75$ instances of the pattern into the graph. We run a search for 100\h000 iterations. And collect the 10 motifs with the best log-factor. We then sample two other graphs by the same method: one with $\bc{k}=0$ and one with $k=150$. We also test each of the 10 motifs found on these two graphs.

 The results are shown in Figure~\ref{figure:synthetic}. For $k=0$, as expected, we find no patterns with positive compression. We also ran a full search on this graph to verify that no motifs can be found unless they are explicitly added to the graph. As in \cite{bloem2017large}, we find that the inserted motif is recovered, even at a low frequency, but many other subgraphs, that share structure properties with the inserted pattern are also marked as motifs. We can recognize the inserted motif as the one, with the highest log-factor, but we see that many of these ``partial motifs'' will be included in the resulting list of patterns with a positive log factor.
 
This experiment only tests a single pattern. To see the effect of multiple random patterns, we repeat the experiment many times, sampling both the pattern and the random graph. 

To sample the pattern we first sample a random number of nodes $n$ from $U(3, 6)$, the uniform distribution over the integer range $(3, 6)$ (including both end points). We then sample a random number of links $m$ from $U(n, n^2 - n)$, and sample a random directed graph from $G(n, m)$. We make $U(0, n)$ nodes and $U(0, m)$ links into variables, choosing constants for the rest uniformly from the data. If the pattern is disconnected, we reject and sample again.

We sample a random graph as in the previous experiments, using the dimensions from the three real world datasets used later. We then add $k$ instances of the motif to the graph and compute the log-factor of the sampled pattern (we do not use simulated annealing here).

We let $k$ range from 0 to 200, and repeat the experiment 25 times for each $k$, sampling a new graph and pattern each time. The results are shown in Figure~\ref{figure:synthrep}. We observe first that under this ad-hoc sampling regime, we produce some patterns that create only very few instances in the graph, after overlapping instances are pruned. Since it is no surprise that these don't allow significant compression, we plot these as small points so that they don't obscure the other points. 

We see that most of the other instances---those that generate enough non-overlapping instances---result in high positive log factors, allowing them to be retrieved as motifs.

\begin{table}[p]
\centering
\tiny
\begin{tabular}{r r p{0.5\linewidth}}
\hline
log-factor & frequency & \\
\hline
\multicolumn{3}{c}{Dogfood , top 5 by log-factor ($>100$ positive)}\\
\hline	
 361495.0 & 10475 & \makecell{\texttt{?n1 dc:creator ?n2.} \\\texttt{?n1 foaf:maker ?n2.} \\\texttt{?n2 foaf:made ?n1.} } \\ 
 \hdashline 244579.5 & 7674 & \makecell{\texttt{?n1 dc:creator ?n2.} \\\texttt{?n1 foaf:maker ?n2.} \\\texttt{?n1 swrs:author ?n2.} } \\ 
  \hdashline 220360.2 & 12138 & \makecell{\texttt{?n1 foaf:maker ?n2.} \\\texttt{?n2 foaf:made ?n1.} } \\ 
 \hdashline  189627.3 & 9888 & \makecell{\texttt{?n1 foaf:member ?n2.} \\\texttt{?n2 swrs:affiliation ?n1.} } \\ 
 \hdashline  187972.9 & 10475 & \makecell{\texttt{?n1 dc:creator ?n2.} \\\texttt{?n2 foaf:made ?n1.} } \\ 
\hline
\multicolumn{3}{c}{Dogfood, top 3 by frequency}\\
\hline	
 -3076.2 & 134853 & \makecell{\texttt{?n1 rdf:\_1 ?n2.} \\\texttt{?n1 rdf:\_2 ?n4.} \\\texttt{?n1 rdf:\_3 ?n3.} } \\ 
 \hdashline  -3435.0 & 116074 & \makecell{\texttt{?n1 swc:heldBy ?n3.} \\\texttt{?n1 swc:heldBy ?n2.} } \\ 
 \hdashline  -2379.9 & 110461 & \makecell{\texttt{?n1 rdf:type owl:Thing.} \\\texttt{?n2 rdf:type owl:Thing.} } \\ 
\hline
\multicolumn{3}{c}{AIFB, top 5 by log-factor ($>100$ positive)}\\
\hline	
 79234.0 & 7549 & \makecell{\texttt{?n1 ?p3 ?n2.} \\\texttt{?n2 ?p4 ?n1.} } \\ 
 \hdashline 61310.4 & 4154 & \makecell{\texttt{?n1 swrs:publication ?n2.} \\\texttt{?n2 ?p3 ?n1.} } \\ 
 \hdashline 57641.1 & 3965 & \makecell{\texttt{?n1 swrs:publication ?n2.} \\\texttt{?n2 swrs:author ?n1.} } \\ 
 \hdashline 57603.1 & 3965 & \makecell{\texttt{?n1 swrs:author ?n2.} \\\texttt{?n2 ?p3 ?n1.} } \\ 
 \hdashline 33168.0 & 7930 & \makecell{\texttt{?n1 swrs:publication ?n2.} \\\texttt{?n2 rdf:type ?n3.} \\\texttt{?n2 swrs:author ?n1.} } \\ 
\hline
\multicolumn{3}{c}{AIFB, top 3 by frequency}\\
\hline
 -908.2 & 181246 & \makecell{\texttt{?n1 swrs:year ?n3.} \\\texttt{?n2 swrs:year ?n3.} } \\ 
 \hdashline -1524.3 & 173059 & \makecell{\texttt{?n1 swrs:publication ?n3.} \\\texttt{?n1 swrs:publication ?n2.} } \\ 
 \hdashline -1667.9 & 103434 & \makecell{\texttt{?n1 swrs:member ?n2.} \\\texttt{?n3 ?p5 ?n1.} \\\texttt{?n4 swrs:author ?n2.} } \\  
\hline
\multicolumn{3}{c}{Mutag, top 5 by log-factor ($87$ positive)}\\
\hline
 178304.4 & 18634 & \makecell{\texttt{?n1 mtg:\_hasAtom ?n3.} \\\texttt{?n1 mtg:\_hasBond ?n2.} \\\texttt{?n2 mtg:\_inBond ?n3.} } \\ 
 \hdashline 97237.8 & 9189 & \makecell{\texttt{?n1 mtg:\_hasAtom ?n2.} \\\texttt{?n2 mtg:\_charge ?n3.} } \\ 
 \hdashline 93819.3 & 8924 & \makecell{\texttt{?n2 rdf:type ?n3.} \\\texttt{?n2 mtg:\_charge ?n1.} } \\ 
 \hdashline 90447.5 & 18634 & \makecell{\texttt{?n1 mtg:\_hasBond ?n2.} \\\texttt{?n2 mtg:\_inBond ?n4.} \\\texttt{?n2 mtg:\_inBond ?n3.} } \\ 
 \hdashline 79027.5 & 8924 & \makecell{\texttt{?n1 mtg:\_hasAtom ?n2.} \\\texttt{?n2 rdf:type ?n3.} } \\ 
\hline
\multicolumn{3}{c}{Mutag, top 3 by frequency}\\
\hline
   -2040.6 & 68514 & \makecell{\texttt{?n1 rdfs:subClassOf ?n2.} \\\texttt{?n3 rdf:type owl:Class.} \\\texttt{?n4 rdf:type owl:Class.} \\\texttt{?n4 rdfs:subClassOf ?n2.} } \\ 
 \hdashline -2077.8 & 60832 & \makecell{\texttt{?n1 ?p5 owl:Class.} \\\texttt{?n3 rdfs:subClassOf ?n2.} \\\texttt{?n4 ?p5 owl:Class.} \\\texttt{?n4 rdfs:subClassOf ?n2.} } \\ 
 \hdashline -1532.6 & 32009 & \makecell{\texttt{?n1 mtg:\_cytogen\_sce "true".} \\\texttt{?n1 mtg:\_salmonella ?n3.} \\\texttt{?n2 mtg:\_amesTestPositive ?n3.} } \\ 
\hline
\end{tabular}
\vspace{2mm}
\caption{Results of the experiment on real-world data. For each experiment we also report the number of motifs found with a positive log-factor.}
\label{table:topk}
\end{table}

\paragraph{Real data} Finally, we will test our method on real data, to confirm that the motifs found coincide with our intuition. We test three datasets: The Semantic Web dogfood dataset \cite{moller2007recipes} ($n=7611, m=242256, r=170$) describing researchers and publications in the Semantic Web domain, the AIFB dataset \cite{bloehdorn2007kernel} ($n=8285, m=29226, r=47$) describing the structure of the AIFB institute, and the Mutag RDF dataset\footnotemark~ ($n=23644, m=74567, r=24$), describing a set of carcinogenic and non-carcinogenic molecules both in structure and properties.

\footnotetext{Originally distributed as an example dataset with the DL-Learner framework \cite{lehmann2009dl}.}

For all datasets, we run 32 parallel searches, with 3125 iterations per search. Table~\ref{table:topk} reports the top 5 motifs by log-factor, and the top 3 motifs by frequency. We provide the top 100 motifs under both criteria online. \footnotemark

\footnotetext{\url{https://github.com/MaestroGraph/motive-rdf}}

The method provides many positives. To see that these are not just random noise, consider those patterns that have high frequency, but a negative log-factor. For instance, the most frequent pattern in the AIFB data describes two entities having the same ``year'' property. Clearly, such a pattern can be matched often, and in many different ways, but it does not provide a satisfying explanation of the the structure of the graph.

Much of what the motif code picks up on is redundancy in the original data. For instance, in the AIFB data both the \texttt{swrs:publication} relation and its inverse \texttt{swrs:author} are always included. Extracting these into a motif is simple way of achieving compression. In fact, the AIFB data contains so many of these relation pairs that the two-node loop with variable labels is the highest scoring motif. In the Dogfood data, we see similar patterns emerge.

Table~\ref{table:selected} shows some interesting motifs from the top 100 for each dataset. We see, for instance that the assertions that something is true or false are both motifs. While these are single triples with only one variable, they occur often enough, that encoding them separately provides a positive compression. The example from the AIFB date shows a typical "star" pattern likely to emerge from relational data: a single entity, surrounded by a set of attributes.

\section{Discussion}

We have presented a new method for mining graph patterns from knowledge graphs. To our knowledge, this is the first method presented that can potentially find arbitrary basic graph patterns to describe the innate structure of a knowledge graph.

\begin{table}[tb]
\centering\small
\begin{tabular}{r r c r}
\hline
log-factor & freq. & &  \\
\hline
220360.2 & 12138 & \makecell{\texttt{?n1 foaf:maker ?n2.} \\\texttt{?n2 foaf:made ?n1.}} & D  \\ \hdashline
 3157.0 & 1011 & \makecell{\texttt{?n1 ?p2 "false".}}  & M \\ \hdashline
  3150.2 & 985 & \makecell{\texttt{?n1 ?p2 "true".}} & M \\ \hdashline
   12871.8 & 8308 & \makecell{\texttt{?n1 rdf:type ?n2.} \\\texttt{?n1 swrs:year ?n3.} \\\texttt{?n4 swrs:publication ?n1.} }& A \\ 
\hdashline
\end{tabular}
\vspace{2mm}
\caption{Selected motifs. The frequency is the number of matches found in the set time limit. The last column indicates the dataset (Dogfood, MUTAG and AIFB, respectively).}
\label{table:selected}
\end{table}

\paragraph{Limitations and future work}

Currently, the greatest limitation of this method is scalability. We note that this limitation only exists when motifs need to be \emph{found}. To \emph{test} whether a given pattern is a motif, the most expensive step required is simply to find instances of the pattern in the graph (as many as is feasible). However, the search space of all patterns is large and complex, making a search for motifs an expensive task.

In \cite{bloem2017large}, the original method on which this method is based was shown to scale to graphs with billions of links. However, this scalability does not translate directly to knowledge graphs: the random walk sampling used there, to generate likely motifs fails in the face of common knowledge graph topologies with many very strong hubs. In such cases, the subgraphs that have a positive log factor are so unlikely to be sampled, that none are ever put to the test. 

For now, we have resorted to black box optimization for search. If a faster search algorithm can be designed specifically for this code, the problem of scaling may be overcome. One option is to replace the random walk used in \cite{bloem2017large} by a biased random walk more suited to the topology of knowledge graphs

Our method currently produces a large number of motifs. We can show that worthwhile motifs are included, and that it performs better than a frequency baseline, but it still takes some manual effort to sort through the suggestions to find the kind of motifs that fit a particular use case. This is not surprising; it is the nature of knowledge graphs that many different and overlapping substructures can be seen as natural or meaningful. One promising avenue to reduce this manual effort is to search for a \emph{set} of motifs which together compress well, each motif claiming a certain part of the knowledge graph to represent. 

\paragraph{Acknowledgements}

This publication was supported by the Amsterdam Academic Alliance Data Science (AAA-DS) Program Award to the UvA and VU Universities. 

\bibliography{kgmotifs}
\bibliographystyle{style/splncs04}

\end{document}


\maketitle 

\tiny
\subsection{Dogfood, top 100 by log factor}

\begin{longtable}{ r r p{10cm} }
 log factor & frequency & \\
\hline\endhead
\input{results/dogfood/motifs-byscore.latex}
 \hline
\end{longtable}
\subsection{Dogfood, top 100 by frequency}

\begin{longtable}{ r r p{10cm} }
 log factor & frequency & \\
\hline\endhead
\input{results/dogfood/motifs-byfreq.latex}
 \hline
\end{longtable}

\subsection{AIFB, top 100 by log factor}

\begin{longtable}{ r r p{10cm} }
 log factor & frequency & \\
\hline\endhead
\input{results/aifb/motifs-byscore.latex}
\hline
\end{longtable}

\subsection{AIFB, top 100 by frequency}

\begin{longtable}{ r r p{10cm} }
 log factor & frequency & \\
\hline\endhead
 \input{results/aifb/motifs-byfreq.latex}
\hline
\end{longtable}

\subsection{Mutag, top 100 by log factor}

\begin{longtable}{ r r p{10cm} }
 log factor & frequency & \\
\hline\endhead
\input{results/mutag/motifs-byscore.latex}
\hline
\end{longtable}

\subsection{Mutag, top 100 by frequency}

\begin{longtable}{ r r p{10cm} }
 log factor & frequency & \\
\hline\endhead
 \input{results/mutag/motifs-byfreq.latex} 
\hline
\end{longtable}